\documentclass[12pt]{article}

\usepackage{scicite}

\usepackage{times}

\usepackage{graphicx}%
\usepackage{multirow}%
\usepackage{amsmath,amssymb,amsfonts}%
\usepackage{amsthm}%
\usepackage{mathrsfs}%
\usepackage[figuresright]{rotating}%
\usepackage[title]{appendix}%
\usepackage{xcolor}%
\usepackage{textcomp}%
\usepackage{manyfoot}%
\usepackage{booktabs}%
\usepackage{algorithm}%
\usepackage{algorithmicx}%
\usepackage{algpseudocode}%
\usepackage{program}%
\usepackage{listings}%
\usepackage{units}
\usepackage{wasysym}

\usepackage{comment}
\usepackage{hyperref}

\hypersetup{
    colorlinks=true,
    linkcolor=black,
    filecolor=black,      
    urlcolor=black,
    citecolor=black,
}

\usepackage[labelfont=bf]{caption}

\usepackage{titlesec}
\titleformat*{\section}{\bfseries\large\uppercase}
\titleformat*{\subsection}{\bfseries\large}
\titleformat*{\subsubsection}{\bfseries\itshape}

\def\citemt{\cite}

\newenvironment{sciabstract}{
\begin{quote} \bf}
{\end{quote}}

\title{Autonomous Medical Needle Steering In Vivo}

\author
{
Alan Kuntz$^{1}$,
Maxwell Emerson$^{2}$,
Tayfun Efe Ertop$^{2}$,
Inbar Fried$^{3}$,
\\
Mengyu Fu$^{3}$,
Janine Hoelscher$^{3}$,
Margaret Rox$^{2}$,
Jason Akulian$^{4}$,
\\
Erin A. Gillaspie$^{5}$,
Yueh Z. Lee$^{6}$, 
Fabien Maldonado$^{5}$,
\\
Robert J. Webster III$^{2}$,
Ron Alterovitz$^{3\ast}$
\\
\normalsize{$^{1}$School of Computing and Robotics Center, University of Utah;}\\
\normalsize{Salt Lake City, UT 84112, USA.}\\
\normalsize{$^{2}$Department of Mechanical Engineering, Vanderbilt University;}\\
\normalsize{Nashville, TN 37235, USA.}\\
\normalsize{$^{3}$Department of Computer Science, University of North Carolina at Chapel Hill;}\\
\normalsize{Chapel Hill, NC 27599, USA.}\\
\normalsize{$^{4}$Department of Medicine, Division of Pulmonary Diseases and Critical Care Medicine,}\\
\normalsize{University of North Carolina School of Medicine; Chapel Hill, NC 27599, USA.}\\
\normalsize{$^{5}$Department of Medicine and Thoracic Surgery, Vanderbilt University Medical Center;}\\
\normalsize{Nashville, TN 37232, USA.}\\
\normalsize{$^{6}$Department of Radiology, University of North Carolina School of Medicine;}\\
\normalsize{Chapel Hill, NC 27599, USA.}\\
\\
\normalsize{$^\ast$Corresponding author. Email: ron@cs.unc.edu.}
}

\date{}

\begin{document} 

\baselineskip24pt

\maketitle

\normalsize{
}

\begin{sciabstract}

\thispagestyle{empty}
Abstract:
The use of needles to access sites within organs is fundamental to many interventional medical procedures both for diagnosis and treatment. Safe and accurate navigation of a needle through living tissue to an intra-tissue target is currently often challenging or infeasible due to the presence of anatomical obstacles in the tissue, high levels of uncertainty, and natural tissue motion (e.g., due to breathing). Medical robots capable of automating needle-based procedures in vivo have the potential to overcome these challenges and enable an enhanced level of patient care and safety. In this paper, we show the first medical robot that autonomously navigates a needle inside living tissue around anatomical obstacles to an intra-tissue target. Our system leverages an aiming device and a laser-patterned highly flexible steerable needle, a type of needle capable of maneuvering along curvilinear trajectories to avoid obstacles. The autonomous robot accounts for anatomical obstacles and uncertainty in living tissue/needle interaction with replanning and control and accounts for respiratory motion by defining safe insertion time windows during the breathing cycle. We apply the system to lung biopsy, which is critical in the diagnosis of lung cancer, the leading cause of cancer-related death in the United States. We demonstrate successful performance of our system in multiple in vivo porcine studies and also demonstrate that our approach leveraging autonomous needle steering outperforms a standard manual clinical technique for lung nodule access. 

\textbf{One-Sentence Summary:} 
We demonstrate the first medical robot that autonomously steers a needle in vivo to an intra-tissue target while navigating around obstacles inside living tissue.

\end{sciabstract}

\vspace{32pt}

\section*{Introduction}\label{sec:intro}

Access to targets inside organs is critical to many interventional medical procedures. 
Such procedures include biopsy, directed drug delivery, ablation, and localized radiation cancer treatment.
These procedures are commonly performed throughout the body in organs such as the lungs, liver, prostate, and brain.
Needles are a fundamental tool used in these procedures as they offer a minimally invasive method for traversing tissue en route to the target site~\cite{Reed2011_RAM, Amack2019, adebar2015methods, swaney2017toward, Minhas2009_EMBS, leipheimer2019first}.
Safe and accurate navigation of a needle through living tissue to an intra-tissue target is currently often challenging or infeasible due to the presence of anatomical obstacles in the tissue, high levels of uncertainty, and natural tissue motion (e.g., due to breathing).
Medical robots capable of automating needle-based procedures in living tissue have the potential to overcome these challenges and enable an enhanced level of patient care and safety.

In this paper, we show the first medical robot that autonomously inserts a needle in vivo to an intra-tissue target while navigating around obstacles inside living tissue. Our system leverages a flexible steerable needle, a type of needle capable of following curvilinear trajectories to avoid obstacles. This property enables steerable needles to reach regions in the anatomy that are otherwise inaccessible using rigid straight instruments~\cite{fried2021design}, making them an appealing tool to overcome many of the challenges faced by existing needle insertion instruments. Precise manual insertion of steerable needles is challenging and unintuitive due to their non-holonomic constraints, making robotic autonomous actuation important~\cite{Majewicz2012_TBE, Majewicz2013_WHC}. Automating a steerable needle procedure in vivo is significantly more challenging than automating an insertion on a benchtop or in ex vivo tissues due to the higher stakes in obstacle avoidance (e.g., puncturing a significant blood vessel can be fatal), the substantial increase in uncertainty in needle/tissue interaction properties, and the need to account for living tissue motion (e.g., due to breathing). Despite the considerable effort put into developing autonomous steerable needles by our group and others~\cite{Patil2010_BioRob, Kuntz2015_IROS, Hoelscher2021_RAL, Fu2021_RSS, Liu2016_RAL, pinzi2019adaptive, favaro2018automatic}, no prior work has reported the autonomous in vivo deployment of a steerable needle to a predefined intra-tissue target in any living organ.

The lung presents one of the most technically challenging applications for autonomous medical robots due to the pervasiveness of obstacles in the tissue and the continuous influence of respiratory motion.
Diagnosing suspicious lesions in the lung parenchyma---the functional tissue of the organ---is an important medical problem since lung cancer, which is the annual leading cause of cancer-related death in the United States accounting for over 130,000 deaths each year, has a drastically more favorable outcome when diagnosed early~\cite{Siegel2021_CA, Howlader2018_NCI}.
The current least invasive way to diagnose lung lesions is via bronchoscopy, where a physician navigates an endoscope through the mouth and into the airways to a target site from which they insert a straight needle into the lung parenchyma.
While existing bronchoscopy techniques can accurately access targets near airways of sufficient diameter to accommodate the bronchoscope, they face accuracy challenges for more distant targets stemming from respiratory motion and from the dense network of airways and vessels that present obstacles to deeper insertions~\cite{yarmus2020prospective, memoli2012meta, ost2016diagnostic}.
This makes existing tools insufficient for diagnosis in many cases.

In this work, we accomplish an intraparenchymal in vivo procedure via steerable needle deployments in one of the most difficult organs---the lungs---overcoming challenges of employing medical robot autonomy inside a living, dynamic organ.
We do this via a bronchoscopically-deployed robotic steerable needle that achieves high curvature through laser patterning and is deployed via an aiming device.
Our system accounts for tissue and respiratory motion by using registration algorithms and defining safe insertion time windows during the breathing cycle during which the needle can safely advance.
It accounts for anatomical obstacles, such as clinically-significant vasculature, bronchi, and the visceral pleural boundary, via motion planning, and it accounts for uncertainty in tissue/needle interaction and intra-operative physician choices with replanning and control. 
The system  includes a user interface that visualizes the status of the procedure and supports autonomous steerable needle deployment.
Prior work on this medical application~\cite{Rox2020_Access, Ertop2020_DSCC, Kuntz2016_Hamlyn} did not consider the challenges of in vivo and
only considered stationary ex vivo tissue without respiratory motion,
targets were selected manually, 
and there was no teleoperated aiming device or re-planning prior to autonomous needle insertion to compensate for human actions and other unmodeled effects. 
There also was no user interface with which the physician could monitor procedure progress, receive cues for correcting their initial needle placement, or monitor the actions of the robotic system.
Through our advancements to enable accurate steerable needle deployment in vivo, we are able to overcome the challenges of existing diagnostic bronchoscopy techniques to reach otherwise hard-to-reach lung targets.

Through in vivo porcine experiments, we show the clinical feasibility of our system and demonstrate the first successful in vivo autonomous needle insertion to a randomly specified intra-tissue target.
We then show our system's ability to outperform an existing clinical approach in accessing peripheral regions of the lung by conducting a study comparing our system to an endobronchial technique used in modern clinical care.
Through this study, we also demonstrate the benefits of automation for the application of endobronchial intraparenchymal needle biopsy.
The autonomous deployment of the steerable needle has the potential to standardize patient care independently of inter- and intra- physician performance, and makes it possible to leverage steerable needles in this clinically-important application.

\section*{Results}

\subsection*{Semi-autonomous steerable needle system overview}

\begin{figure*}[b!]
    \centering
    \makebox[\textwidth][c]{\includegraphics[width=0.955\textwidth]{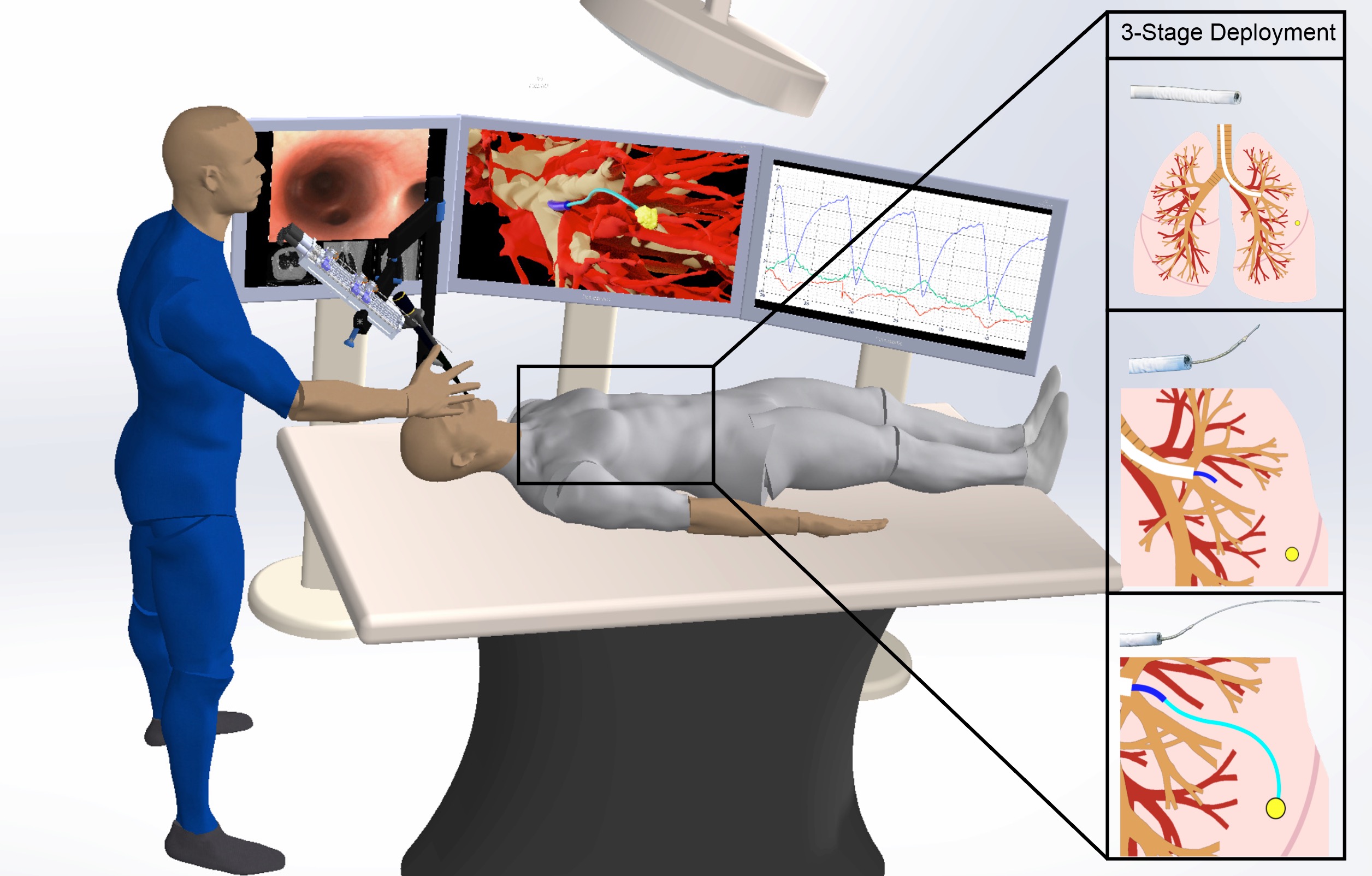}}
    \caption{
    \textbf{Overview of the semi-autonomous medical robot's three stages in the lungs.}
    Stage 1 consists of a physician manually navigating a traditional bronchoscope through the airways. Stage 2 consists of a physician teleoperating the aiming device. Stage 3 consists of autonomous needle deployment through the parenchyma to the nodule while accounting for respiratory motion and avoiding anatomical obstacles such as clinically-significant vasculature, bronchi, and the visceral pleural boundary. 
    }
    \label{fig:fig1}
\end{figure*}

Our system consists of three stages, each of which corresponds directly to a portion of the medical procedure (see Fig.~\ref{fig:fig1}).
To inform each stage, we segment the anatomical environment and obstacles from a preoperative CT scan of the patient.
Stage 1 consists of a traditional bronchoscope, augmented with a 6-DOF electromagnetic tracking coil attached to its distal tip, inserted by the physician through the mouth of the patient and into the airways. 
The physician navigates the bronchoscope inside the airways to a position from which subsequent stages of the system can be deployed to reach the target. 
This portion of the procedure is performed manually, leveraging the experience of the physician in bronchoscope navigation.
Stage 2 consists of a teleoperated
aiming device that the physician passes through the working channel of the bronchoscope following navigation.
The aiming device is a tendon-actuated beveled nitinol tube laser-patterned to increase its 
achievable levels of curvature
~\cite{Rox2020_ICAIM}.
The aiming device enables the physician to reorient 
the steerable needle with respect to the target in the parenchyma prior to autonomous needle deployment.
This is an important step in successfully transferring control from the physician (who navigated the bronchoscope) to the robot (that will autonomously steer the needle to the target) as it enables the physician to correct deviations in the current needle pose from the planned start pose.
Stage 3 consists of a steerable needle~\cite{Rox2020_Access} which we pass through the working channel of the bronchoscope and through the aiming device. 
The needle is highly flexible and features a bevel tip, which causes it to curve in the direction of the bevel when inserted into parenchymal tissue. The needle's direction of curving can be changed by reorienting its bevel tip by axially twisting the needle at its proximal end. 
We laser-patterned the needle to achieve higher levels of curvature while minimizing the needle diameter~\cite{Rox2020_Access}.
We embedded a 5-DOF tracking coil within the needle at its distal tip.
The robot autonomously deploys the needle along a planned trajectory under closed-loop control to the target while avoiding obstacles~\cite{Hoelscher2021_RAL,Ertop2020_DSCC}.
The robot steers the needle autonomously to overcome the known challenges of manual deployment of steerable needles~\cite{Majewicz2012_TBE, Majewicz2013_WHC}.
The three stages' control methods---manual, tele-operation, and autonomous---harness the respective strengths of the human and robot, with bronchoscope navigation relying on the physician's existing expertise, the tele-operated aiming device serving as a bridge between manual control and robotic automation, and the final stage leveraging the robot's capabilities to autonomously maneuver the steerable needle accurately to the target.

\subsection*{Workflow of the system deployment}

\begin{figure*}[b!]
    \centering
    \makebox[\textwidth][c]{\includegraphics[width=0.85\textwidth]{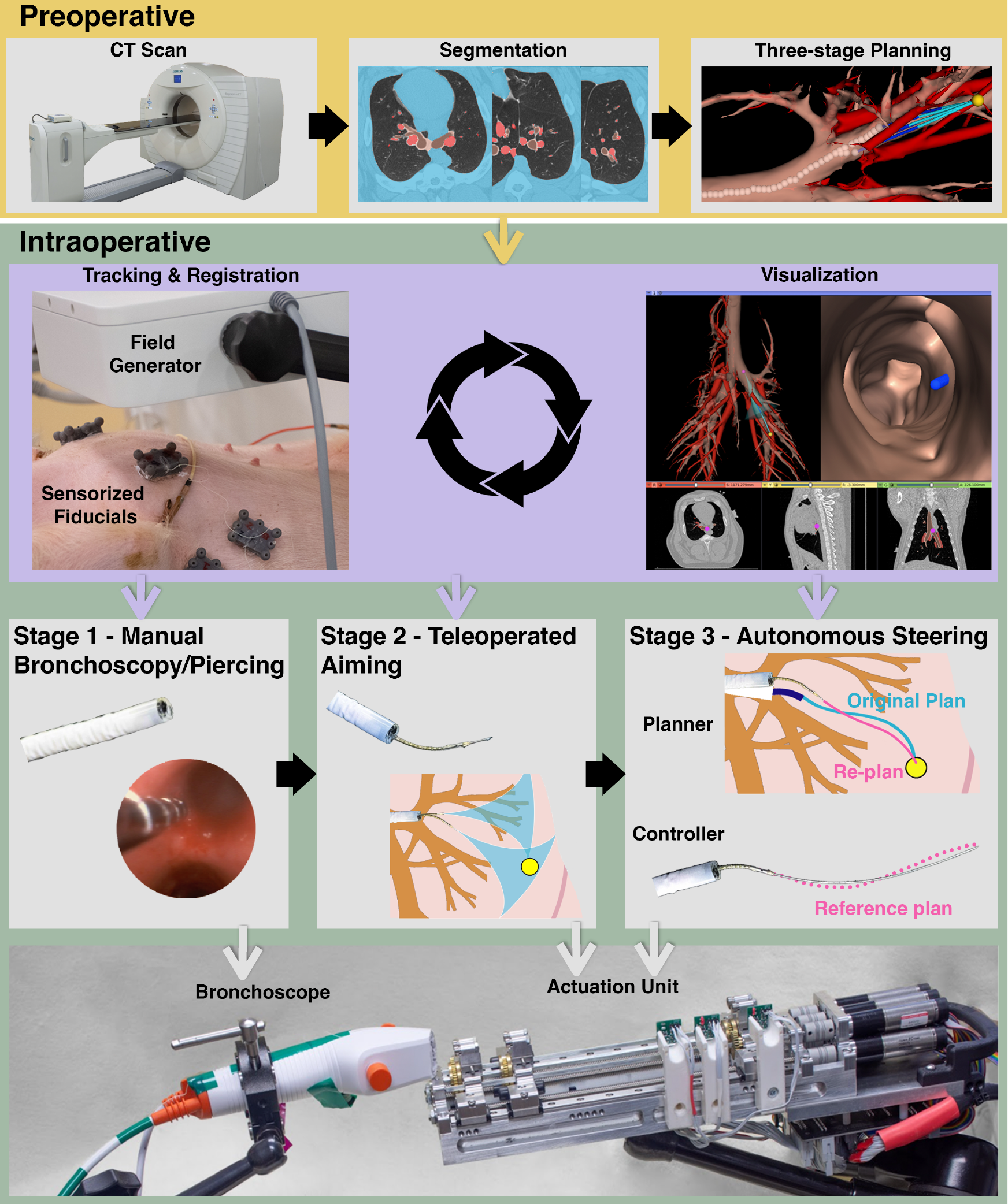}}
    \caption{
    \textbf{Workflow of the procedure showing the integration of the software and hardware components.}
    We show both the procedure sequence flow (thick black arrows) and information flow (thin colored arrows).
    A CT scan is segmented and used to construct a map leveraged by the 3-stage motion planning software in the preoperative phase. The scan, segmentation, and 3-stage plan are used intraoperatively in the procedure, during which the devices are tracked, registered to the CT-frame, and visualized for the operator. These steps inform the 3-stage deployment of the system and the interaction between the physician and the robot.
    }
    \label{fig:fig2}
\end{figure*}

Our system enables accurately accessing lung lesions. The workflow of the process, as shown in Fig.~\ref{fig:fig2}, consists of creating a geometric model of the patient's anatomy, planning the deployment of each of the three stages via motion planning, and finally executing the plan via manual control, teleoperation, and automation.

In the pre-operative phase, we acquire a CT scan of the patient's chest under a breath hold.
The breath hold, commonly utilized for imaging and other clinical procedures such as targeted radiation therapy~\cite{Kubo1996_PMB}, performs two important functions.
First, it reduces motion artifact in the scan from natural tissue motion, and second, it captures a state of the environment that we can accurately return to with subsequent breath holds.
We then segment the CT scan using an automatic segmentation algorithm~\cite{Fu2018_IROS}.
This provides us with a three-dimensional representation of the underlying patient anatomy, including the bronchial tree, blood vessels, and pleural boundary.
We use the segmentations to specify the planning environment for the motion planner, including the free workspace and obstacles.
We also segment the target nodule. In our in vivo experiments, a target is randomly selected at this stage to mimic a clinical procedure. 
For a given target
in the lung parenchyma, we use our motion planning algorithm to generate 3 - 5 candidate plans, each plan composed of a motion plan for each of the three stages~\cite{Hoelscher2021_RAL}.
Each plan has a cost that considers proximity to obstacles.
The candidate plans are overlaid in the segmented anatomy and displayed to the physician in the user interface.
The physician then chooses a plan.

Prior to bronchoscope navigation, we perform registration using sensorized fiducials that are attached to the exterior chest wall of the patient.
We improve this registration by driving the bronchoscope through the airways and collecting a point-cloud of the airway's medial axis in the electromagnetic tracker frame.
We perform a point-cloud registration between the set of collected points and the skeleton of the segmented bronchial tree using the iterative closest point (ICP) algorithm~\cite{Besl1992_TPAMI}.
This additional registration step is particularly important in vivo as it helps account for internal tissue deformation that the fiducials on the chest wall, given their distance from the lung, are not able to accurately capture.
Throughout the procedure, the poses (positions and orientations) of the bronchoscope and steerable needle are tracked in real-time using an electromagnetic (EM) tracking system (Aurora, Northern Digital, Inc.).
The poses are overlaid in the anatomy using the registration transformation and displayed to the physician via the user interface.

Next the physician begins the system deployment.
The physician first manually navigates the bronchoscope based on the three-stage plan they selected.
Next the physician pierces into the lung parenchyma using a piercing stylet deployed through the working channel of the bronchoscope.
The physician then inserts the aiming device through the working channel, passed into the parenchyma over the piercing stylet.
The piercing stylet is then removed and replaced with the steerable needle.
If any error occurred in the manual deployment that resulted in deviation from the planned trajectory, either from operator error or tissue deformation, the system directs the physician to adjust the position and orientation of the steerable needle to best align with the target.
This is done by teleoperating the aiming device.
The system automatically indicates to the physician in the user interface when the realignment is successful.
Once the physician reorients the steerable needle toward the target, the system recomputes an obstacle-free motion plan for the needle in the parenchyma to the target, accounting for the current start pose of the needle.
The robot then, over a sequence of several breath holds, autonomously steers the needle along the planned trajectory to the desired target.

\subsection*{Needle steering under respiratory motion}

\begin{figure*}[b!]
    \centering
    \makebox[\textwidth][c]{\includegraphics[width=\textwidth]{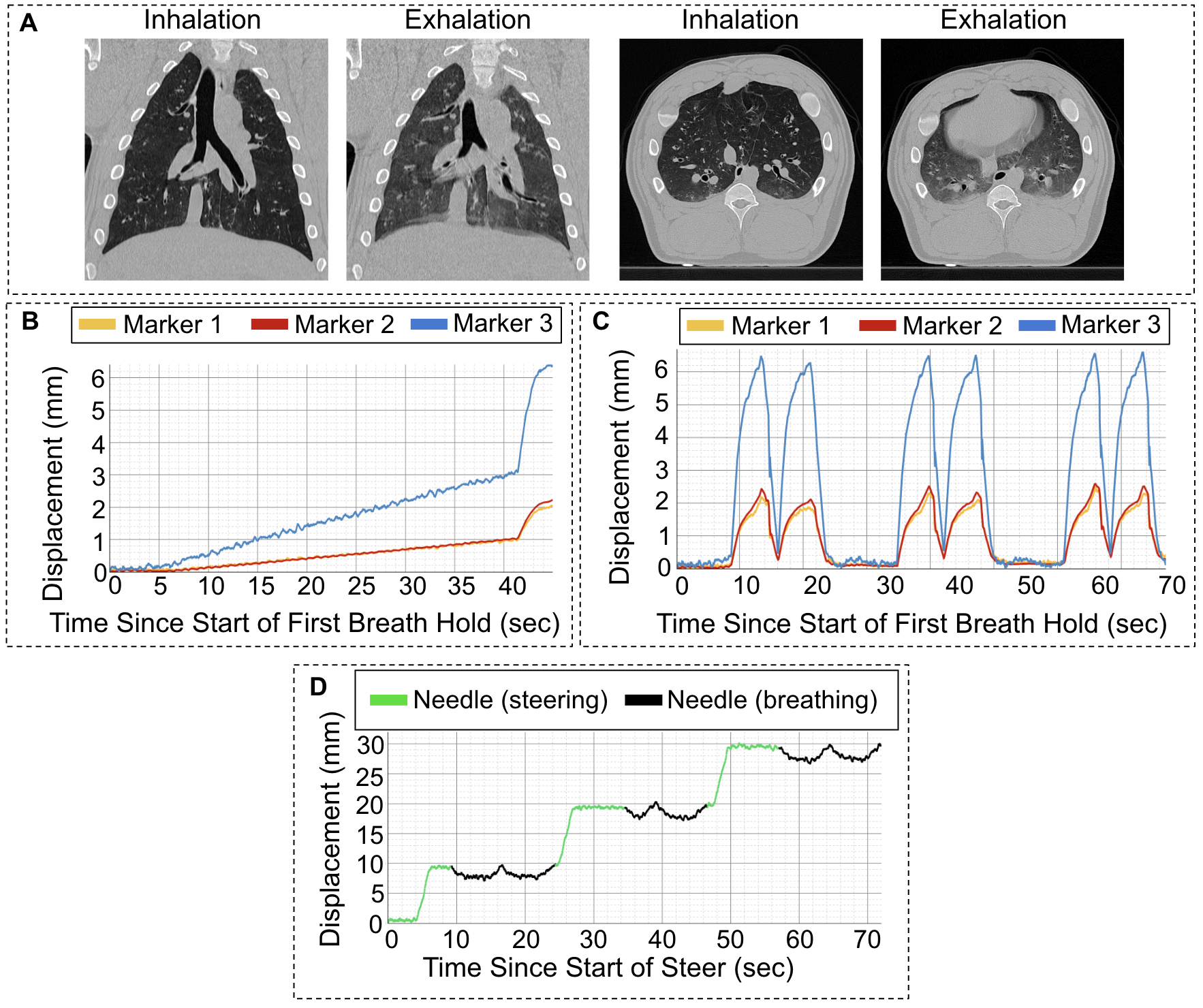}}
    \caption{
    \textbf{Effect of respiratory motion on the procedure.}
    (\textbf{A}) Transverse and coronal views from CT scans taken at peak tidal volume and at functional residual capacity (inhalation and exhalation, respectively).
    (\textbf{B}) Chest marker displacement over a single long breath hold (breath hold stops at approximately 40 seconds).
    (\textbf{C}) Chest marker displacement over a sequence of short breath holds, where the displacements remain under 1mm during approximately 10-second breath holds and rise substantially only between the breath holds.
    (\textbf{D}) Displacement of the needle during one of the in vivo deployments.
    }
    \label{fig:fig3}
\end{figure*}

The lungs deform significantly during breathing, displacing the parenchyma and the anatomy embedded within.
This leads to a discrepancy between the preoperative segmentation of the anatomy and the intraoperative state of the anatomy.
To account for this change in the environment and achieve in vivo needle steering in the lung while avoiding obstacles, we define safe insertion time windows during the breathing cycle
when the needle can safely advance. Specifically, we monitor the respiratory cycle using the fiducials attached to the exterior chest wall by plotting the displacement of the fiducials over time. 
We define the start of the safe insertion time window as the phase of the breathing cycle corresponding to peak tidal volume, i.e., a state of full inhalation. 
We select this as the start of the time window for three reasons. 
First, the CT scan used to segment anatomy is typically acquired in a state of full inhalation, which provides the most detailed view of the target and anatomical obstacles. 
Second, compared with other phases of the breathing cycle, during peak tidal volume the lung is most expanded, meaning that there is more free space for steering the needle (Fig.~\ref{fig:fig3}A).
Third, peak tidal volume represents a discrete state of the lung that is easy to repeat.
At the start of the safe insertion time window, we initiate a breath hold and the robot begins steering the needle. After a brief breath hold, the safe insertion time window ends and the robot pauses insertion of the needle. 
The breathing cycle then continues normally without the robot actuating the needle, and we require at least two normal breathing cycles before initiating another breath hold. 
Since insertion of the steerable needle requires more time than a single safe insertion time window allows, we repeat the process each time the breathing cycle re-enters a safe insertion time window until the needle has reached the target. 

To define the duration of the safe insertion time window, we studied the impact of varying the duration during an in vivo study. 
A series of short breath holds is preferable to a long breath hold for the overall physiologic stability of the animal, and also is beneficial for reducing undesirable tissue motion.
The ability of the animal and ventilator to maintain a breath hold decreases over time, leading to a steady deflation.
This can be seen by monitoring the displacement of the chest markers over a long, approximately 40 second, breath hold (Fig.~\ref{fig:fig3}B).
Performing a sequence of short, approximately 10 second, breath holds reduces the magnitude of the deflation (Fig.~\ref{fig:fig3}C).
By only requiring short breath holds, we are better able to ensure that any needle motion in the tissue is planned and caused by the robot and not a result of natural tissue motion.
Combined with the delay in deflation, this enables us to achieve a consistent lung state for the duration of each segment of steering and globally across segments for the entire steer.
Based on a 10-second safe insertion time window, our system divides the planned needle trajectory into sequential trajectories with arclengths of \unit[10]{mm} and the beginning of each of these trajectory segments is synchronized with the beginning of a breath hold (Fig.~\ref{fig:fig3}D).
The robot then inserts another trajectory segment in each safe insertion time window until the trajectory is completely executed.

\subsection*{In vivo intraparenchymal needle steering in lung}

\begin{figure*}[t!]
    \centering
    \makebox[\textwidth][c]{\includegraphics[width=\textwidth]{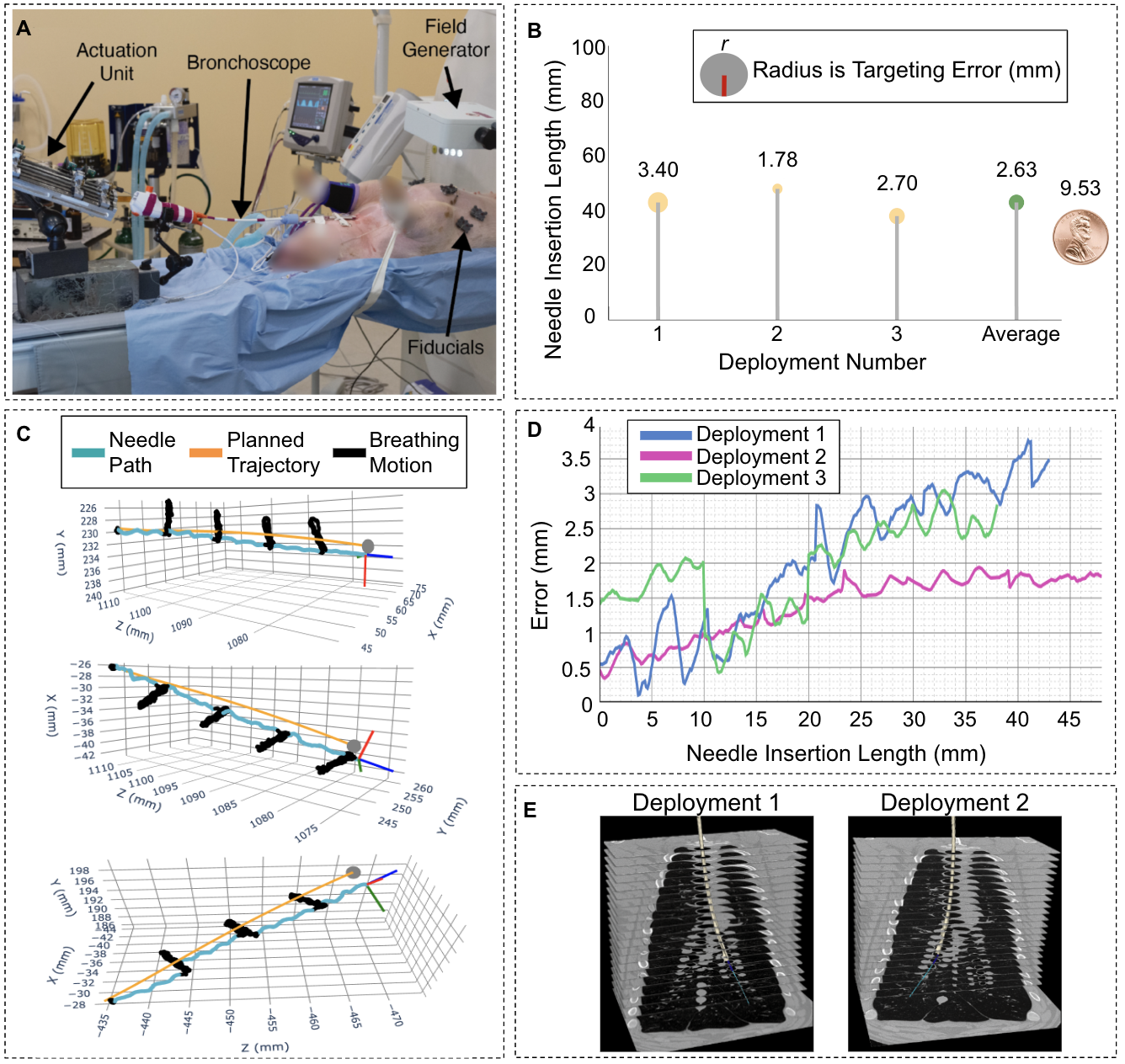}}
    \caption{
    \textbf{In vivo porcine lung results.}
    (\textbf{A}) The in vivo experimental setup.
    (\textbf{B}) Results for all system deployments with needle steer length on the y-axis, steer number on the x-axis, and a circle with radius equal to the targeting error. The penny is for scale.
    (\textbf{C}) Planned and tracked needle trajectories.
    (\textbf{D}) Magnitude of the trajectory error for each deployment.
    (\textbf{E}) Visualization of the deployments shown within slices of the CT scan of the anatomy.
    }
    \label{fig:fig4}
\end{figure*}

Prior work by us and others has demonstrated ex vivo needle steering in lung tissue, as well as other organs~\cite{Rox2020_Access, Reed2011_RAM, adebar2015methods, Ertop2020_DSCC}.
These experiments are important for characterizing the properties and capabilities of the devices, but are insufficient to indicate their clinical feasibility.
This is due to the major differences between in vivo and ex vivo environments including respiratory motion, tangible medical risks, and higher levels of uncertainty in the procedure.
These differences are especially prominent in the lungs---an organ under constant physiological motion.
Leveraging a porcine model, commonly used in research due to its similarity in size and structure to human chest anatomy~\cite{Judge2014_JRCMB}, we assess our system's performance in a realistic setting and demonstrate its clinical potential via in vivo porcine experiments (Fig.~\ref{fig:fig4}).

We performed three deployments of our system in two live
animals (\unit[102]{kg} and \unit[42]{kg}).
Prior to each system deployment, we randomly sampled a target in the lung parenchyma from the initial CT scan that was used for segmenting the anatomy.
The final error for each steer is calculated in the EM tracker space and is measured using the Euclidean distance between the tracking coil at the needle tip and the target.
The first procedure was performed in the left lung lobe of the first animal with an overall steerable needle trajectory length of \unit[43]{mm}.
The final error at the target was \unit[3.4]{mm}.
The second procedure was performed in the right lung lobe of the first animal with an overall steerable needle trajectory length of \unit[48]{mm}.
The final error at the target was \unit[1.8]{mm}.
The third procedure was performed in the left lung lobe of the second animal with an overall steerable needle trajectory length of \unit[38]{mm} and with a final targeting error of \unit[2.7]{mm}.
The quantitative metrics of the three deployments are visualized in Fig.~\ref{fig:fig4}B, with a penny as a scale reference to better illustrate the small magnitude of the targeting errors.
The planned and tracked trajectories of the three steers are shown in Fig.~\ref{fig:fig4}C.
The black segments in the tracked points represent the needle motion during normal tidal breathing in between sequential breath holds.
These plots demonstrate that the needle was able to return to the same position 
over each period of tidal breathing between the sequence of short breath holds.
The success of the system for the three steers is further depicted in Fig.~\ref{fig:fig4}D, showing the magnitude of the trajectory error over time where trajectory error is measured as the Euclidean difference between paired points in the tracked and planned needle trajectory.
The targeting errors of the three deployments in vivo are consistent in magnitude with the targeting errors achieved in ex vivo system deployments during the comparison study described below.

In all procedures, the autonomous steerable needle did not collide with any anatomical obstacles.
Confirmed via post-deployment CT scans, notably there were no adverse events during any of the system deployments, including pneumothorax, atelectasis, or vessel perforation and hemorrhaging.

\subsection*{Comparison with manual clinical bronchoscopy}

\begin{figure*}[b!]
    \centering
    \makebox[\textwidth][c]{\includegraphics[width=\textwidth]{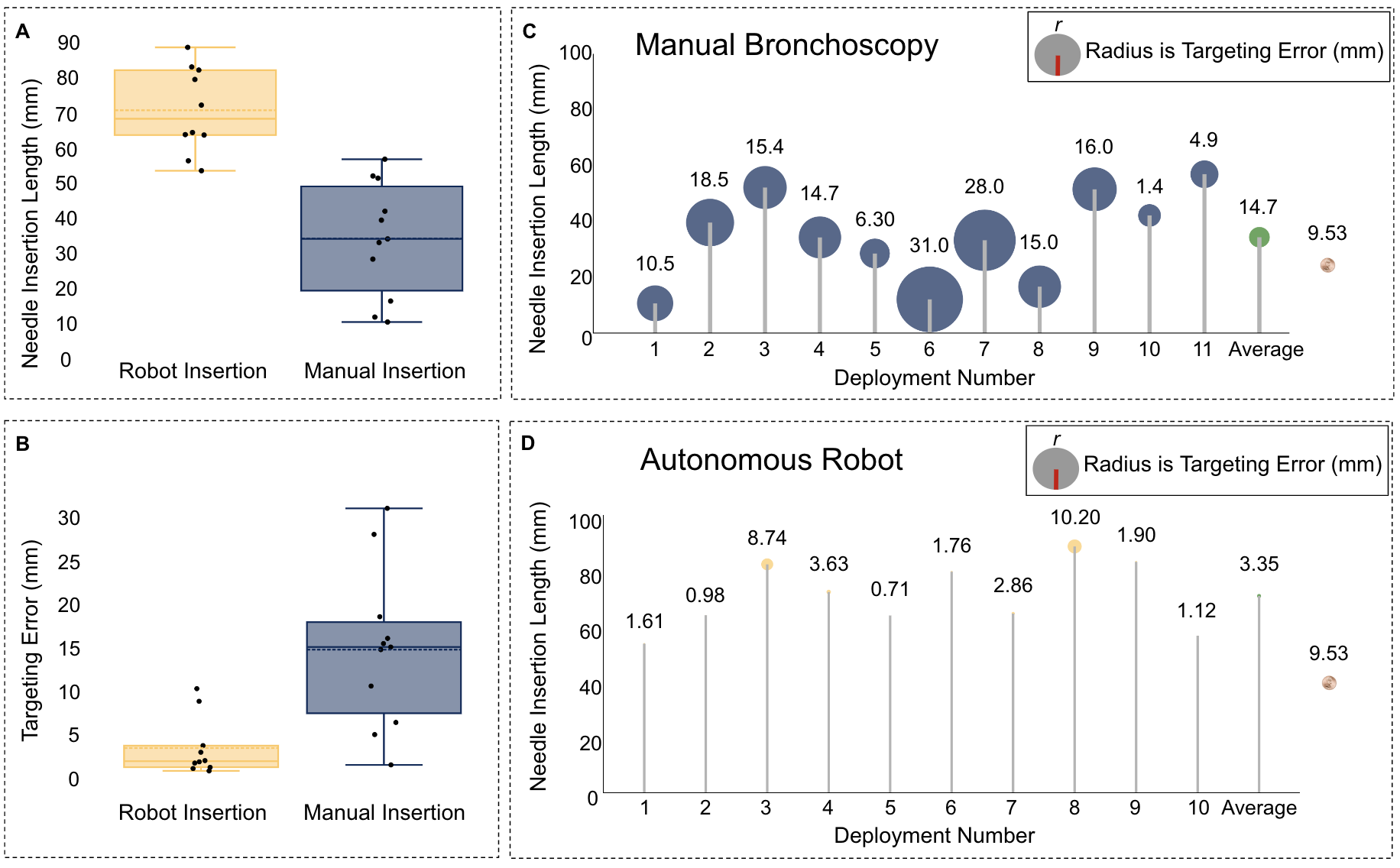}}
    \caption{
    \textbf{Results from the comparison experiments between the manual bronchoscopy and the proposed robotic system.}
    (\textbf{A}) Deployment lengths.
    (\textbf{B}) Targeting errors.
    (\textbf{C}) Results from the manual bronchoscopy deployments.
    (\textbf{D}) Results from the robotic steerable needle deployments.
    }
    \label{fig:fig5}
\end{figure*}

To evaluate our system's ability to improve access to hard-to-reach peripheral targets, we conducted a study in ex vivo porcine lung comparing our system to a modern diagnostic bronchoscopy technique.
For the manual diagnostic bronchoscopy, two expert interventional pulmonologists
navigated a clinical bronchoscope (Ambu) with electromagnetic navigation and CT guidance.
Electromagnetic navigation was provided by embedding an electromagnetic sensor at the distal tip of the bronchoscope and visualizing the real-time pose of the bronchoscope in our visualization software registered to the segmented anatomy.
The study was performed in a CT scanner room, which provided the physicians the opportunity to utilize the scanner for localization confirmation throughout the procedure.
Following navigation of the bronchoscope, the physician inserted a rigid straight needle with an electromagnetic sensor embedded at its tip toward the target.
The stiffness of the rigid needle was consistent with stiffness properties of existing clinical biopsy tools used in bronchoscopy.

For the robotic system, an engineer experienced with the system performed the deployments.
The engineer was provided the same electromagnetic navigation and visualization during the bronchoscope navigation as was given to the physicians in the clinical bronchoscopy cases.

To assess the ability of each approach to accurately access peripheral targets, we randomly sampled virtual targets in the segmented anatomy from peripheral regions of the ex vivo lungs.
The measured outcomes included the insertion length of the devices in the lung parenchyma and the targeting error.
Targeting error was measured as the Euclidean distance between the tracked needle tip and the target.

\begin{figure*}[b!]
    \centering
    \makebox[\textwidth][c]{\includegraphics[width=\textwidth]{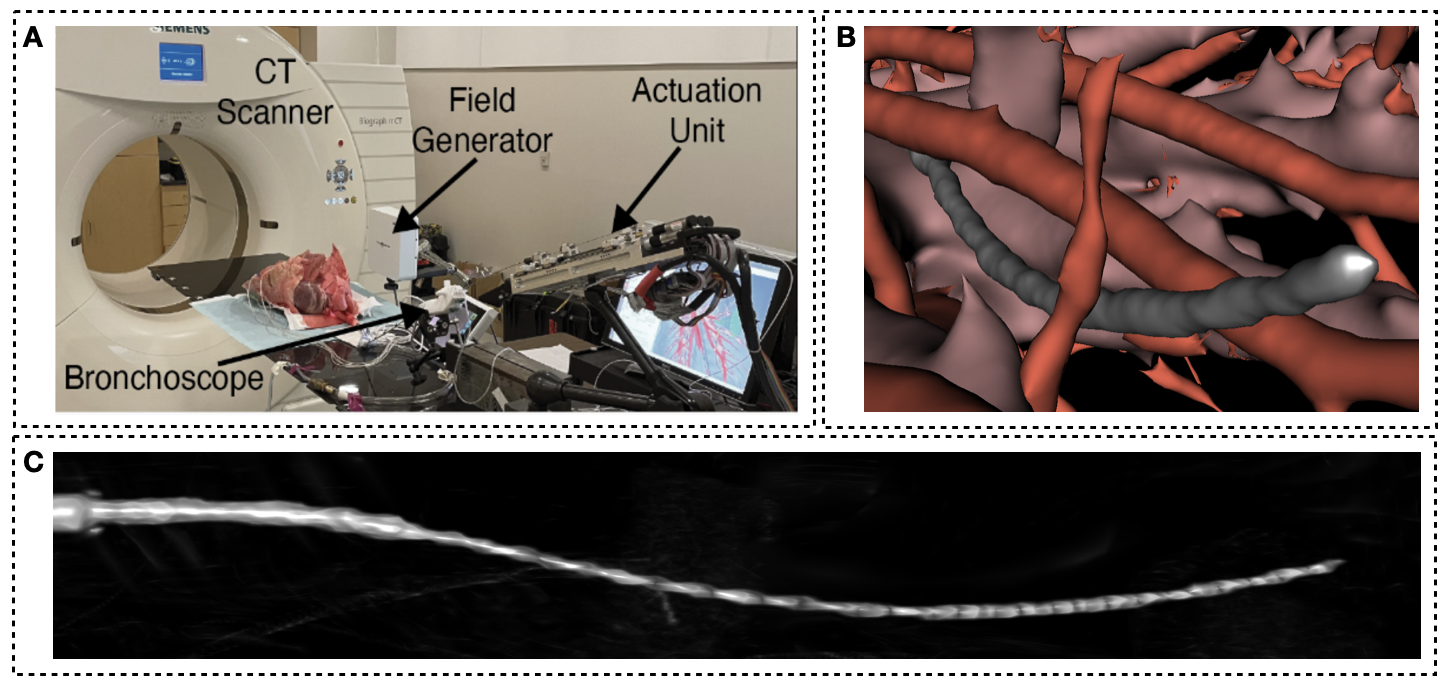}}
    \caption{
    \textbf{Capabilities of the flexible steerable needle.}
    (\textbf{A}) The ex vivo experimental setup. 
    (\textbf{B}) Segmentation of the steerable needle following deployment in ex vivo lungs. 
    (\textbf{C}) The 3D rendering of the steerable needle corresponding to the segmentations.
    }
    \label{fig:fig6}
\end{figure*}

We performed 11 deployments of the clinical bronchoscopy in 5 unique lungs, and 10 deployments of the robot system in 6 unique lungs.
The distribution of the needle insertion lengths and the targeting errors for our system and the clinical bronchoscopy are shown in Fig.~\ref{fig:fig5}A and Fig.~\ref{fig:fig5}B, respectively.
The average needle insertion length when using the robot system was longer than the average needle insertion length when performing the clinical bronchoscopy (\unit[70.8]{mm} $\pm$ \unit[11.5]{mm} and \unit[34.2]{mm} $\pm$ \unit[15.4]{mm}, respectively, \textit{p} = \unit[0.000013]).
The targeting error for the robot deployments was substantially lower than that for the clinical bronchoscopy (\unit[3.4]{mm} $\pm$ \unit[3.2]{mm} and \unit[14.7]{mm} $\pm$ \unit[8.6]{mm}, respectively, \textit{p} = \unit[0.0014]).
The needle insertion length and targeting error of each individual deployment are shown in Fig.~\ref{fig:fig5}C, with a penny as a reference scale for the magnitude of the targeting errors.
The ability of the steerable needle to achieve curvilinear trajectories and avoid anatomical obstacles is shown in Fig.~\ref{fig:fig6}.

During the experiments, we observed that the physician preferred to drive the bronchoscope further into the airway to enter the parenchyma at a more distal point compared with the more proximal insertion recommendations computed by the motion planner for the robot.
The physician's intention was to enter the parenchyma at a point closest to the target, but given the decreasing diameter of the airways, this restricted the range for angling the bronchoscope's distal tip.
By exiting the airway at a more proximal site, our system provided more actuation and better orientation positioning for the bronchoscope navigation.
We also noticed that the physician-guided needle insertions were on average of greater length than existing clinical practice, likely reflecting the high risk tolerance of the physician in a non-clinical setting.
Since manual needle insertion is not explicitly planned with respect to obstacles in the environment in current clinical practice, most clinical tools are limited to \unit[20 - 30]{mm} of insertion~\cite{puchalski2021robotic, liu2015evolution}.
Our system is able to access more peripheral targets via autonomous needle steering by explicitly computing an obstacle-free trajectory in the parenchyma and following it under closed-loop robotic control.

These results demonstrate our system's ability to reach peripheral intraparenchymal targets in the lung, which are challenging for existing bronchoscopic techniques, with high accuracy.

\section*{Discussion}

In this work, we demonstrated autonomous intraparenchymal needle steering to a target in the highly challenging and uncertain environment of in vivo lungs.
We achieved this using a bronchoscopically-deployed robotic steerable needle that achieves high curvature through laser patterning and is
deployed via an aiming device. We accounted for tissue and respiratory motion by using registration algorithms and defining safe insertion time windows during the breathing cycle during which the needle can safely advance. We accounted for anatomical obstacles via motion planning and accounted for uncertainty in tissue/needle interaction and intra-operative physician choices with replanning and control. The system's user interface displays real-time tracking in a segmented view of the anatomy and supports autonomous steerable needle deployment. 
To the best of our knowledge, no prior autonomous medical robot has completed a task in an obstacle-cluttered parenchyma of an organ where visual supervision is not possible and an implicit roadmap of the anatomy, such as the lumen of a blood vessel, does not exist~\cite{Fagogenis2019_SR, Chen2020_NMI, Zevallos2021_ISMR, saeidi2022autonomous, kim2022telerobotic, leipheimer2019first}.
Prior work on needle steering in vivo focused on measuring needle/tissue interaction properties or tele-operation without consideration of patient-specific obstacles \cite{Majewicz2012_TBE,Secoli2022_PLOSOne}.
We demonstrated the clinical feasibility of our semi-autonomous system through in vivo porcine experiments and through a comparative study where our system outperformed a modern clinical approach in safely and accurately accessing hard-to-reach peripheral intraprenchymal targets.
The results of the comparative study also highlight the benefits of automation for lung nodule biopsy, enabling our system to leverage steerable needles to safely traverse the lung parenchyma.
Automating medical procedures and subtasks
has the potential to standardize patient care independently of the natural variation in inter-physician and intra-physician performance while also refocusing the time of expert physicians to higher level tasks.

The results of our experiments indicate that steerable needles could enable physicians to access targets in the lung for diagnosis or treatment that are beyond the safe reach of existing tools.
By integrating our robotic system with a conventional bronchoscope, we limit the additional training required to operate our system, which will facilitate its availability to patients without access to highly sub-specialized physicians or tertiary care centers.
The decoupling of the robotic steerable needle system from the bronchoscope makes it so that our system could be used with any bronchoscopic tool with a sufficient working channel.

The system and results presented in this work highlight the clinical potential of steerable needles, enabling physicians to safely access regions in living soft tissue while avoiding obstacles.
Given our success in the highly challenging environment of the lung, we envision that the third stage of our system, the autonomous steerable needle robot, could be useful in other clinical applications where steerable needles have been proposed, including drug delivery, brachytherapy, ablation, and biopsy in the brain, prostate, and liver~\cite{Abolhassani2007_MEP, adebar2015methods, Minhas2009_EMBS, seifabadi2012design}.

Despite the success of the system, there are some limitations and areas for development before clinical application.
Motivated by the benefits of automating the steerable needle as shown through the comparative study, we plan to automate the teleoperated aiming device in future work to autonomously correct for deviations in the piercing location from the planned exit point.
Also, although the porcine model is a great translational model for respiratory medicine, the anatomy of the porcine lung differs in a couple significant ways from the anatomy of human lungs.
These differences include a monopodial branching structure of the airways compared to the bipodial structure of human airways, large blood vessels that run adjacent to the main airway branch bilaterally making it more challenging to pierce into the parenchyma in porcine lungs, and an absence of the pores of Kohn which prevent collateral ventilation in porcine lungs and result in more atelectasis~\cite{Judge2014_JRCMB, azad2016geometric}.
All of these anatomical differences make the porcine anatomy a more challenging procedural environment than human anatomy, and the next step in this research is to assess the performance of our system in a human cadaver followed by live human trials.

\section*{Materials and Methods}\label{sec:Method}

The goal of this work was to demonstrate the clinical capabilities of autonomous medical robots to reach targets inside the parenchyma of a living organ.
To accomplish this goal, we developed a semi-autonomous steerable needle robotic system and designed in vivo and ex vivo experiments to assess its capabilities.
We chose the lungs as our clinical application because of its clinical significance and because it represents one of the most challenging environments for intraparenchymal medical robots.

\subsection*{Steerable needle robotic platform}

The needle actuation unit consists of three carriages: two for independently controlling translation and rotation of the aiming device and the flexure-tip steerable needle, as well as a third carriage for actuating the pull-wire of the aiming device.
These carriages travel along lead screws driven by motors at the base of the unit that are controlled by custom electronics boards.
The system features a quick-loading tube carriage concept, where each tube in the system has a driven gear that snaps into place via holding arms and engages with a driving gear on the carriage, enabling the physician to easily switch between tools if necessary~\cite{Amack2019}.

\subsection*{Fiducial Registration}

To register between the electromagnetic field generator frame and the CT scanner frame, we utilize 3D printed fiducials that are placed on the outer surface of the lungs in ex vivo experiments and on the chest wall for in vivo experiments.
The fiducials are rigid plastic markers with a geometry that includes seven spheres.
In each fiducial we embed a 6 DOF electromagnetic tracking coil.
Each fiducial is pre-calibrated prior to the experiment such that the center of each sphere is known relative to the embedded tracking coil.
This is done by placing a probe in the center of each sphere (into a divot designed to the size of the probe tip) while the location of the central tracker is also recorded. 
The spheres of every fiducial are easily recognizable in CT image space.
As such, we manually segment the location of the sphere centers for each fiducial from the preoperative CT scan.
Then, given the known point-to-point correspondence between the EM tracked sphere centers and the sphere centers in the CT scan, we utilize a registration algorithm~\cite{Fitzpatrick2000_HMI} to line up the CT image space with the magnetic tracking space.

\subsection*{Tree-to-Tree Registration}

To improve the registration, we refine the fiducial registration using the correspondence-free iterative closest point (ICP) registration method \citemt{Besl1992_TPAMI}.
We perform this registration between a set of EM-sensed internal airway points and the medial axis of the airway segmented from the CT scan.
To collect the EM-sensed points, we deploy a custom tool composed of a plastic sheath with an embedded 6 DOF tracking coil through the bronchoscope tool channel.
Prior to executing the planned intervention, the physician first steers the bronchoscope and this tracking tool through a large number of the patient's airways.
This process enables us to collect a large point cloud in the EM-tracker space representing the airways.
This point cloud is then registered with the medial axes of the airway segmentation from the CT scan.
We combine these two approaches by using the fiducial registration as the initialization of the ICP method.

\subsection*{In vivo experimental setup}

The in vivo porcine experiments were conducted at the University of North Carolina (UNC) at Chapel Hill
under the supervision of a veterinary team from the UNC Division of Comparative Medicine using a protocol approved by the Institutional Animal Care and Use Committee (IACUC).
The veterinary team anaesthetized the animals (\unit[102]{kg}, \unit[42]{kg}) using a solution of Telazol, Ketamine, and Xylazine, all at \unit[1]{ml/kg}.
They used inhaled isofluorane for maintenance of anesthesia.
The veterinary team performed a tracheotomy to circumvent the porcine upper respiratory anatomy which differs from that of humans and is often difficult to intubate~\cite{Judge2014_JRCMB}. 
The animals were placed on a volume-controlled bellows ventilator (Hallowell EMC Model 2000).
We used a positive end-expiratory pressure (PEEP) valve (\unit[10]{cm H$_{2}$O}) to prevent lung collapse in the peripheral airways.
To minimize residual breathing effort and negate the involuntary coughing reflex, the veterinary team administered paralytic (Rocuronium Bromide \unit[10]{mg/mL}) via a syringe pump (Medfusion 2010i) at an infusion rate of \unit[10]{mL/hr}.
We acquired a CT scan of the lungs using a Siemens Biograph mCT with 140 kVp, 0.8 pitch, and \unit[1.0]{mm} slice thickness.
The CT scan parameters are similar to those used in a clinical chest CT for procedure planning.
The initial CT scan was performed during a breath hold in order to stabilize the anatomy to capture a reference image.
 
Following segmentation, we randomly sampled targets in the porcine lung representing suspicious nodules for the device to reach.
The targets were sampled from manually-placed cuboidal regions corresponding to human-like anatomical regions of the porcine lung, as informed by interventional pulmonologists on our team.
The deployment stages were then carried out, as described in the main body of the paper.
At the end of the procedure
the animal was euthanized by the veterinary team.

\subsection*{Ex vivo experimental setup}

The ex vivo experiments were conducted at UNC-Chapel Hill.
We intubated the lungs by inserting an endotracheal tube through the trachea and inflated the lungs with a constant airway pressure.
The pressure was initially set to \unit[1]{psi} and adjusted for each lung to compensate for air leaks caused by minor damage to the lung suffered when it was harvested.
We attached fiducial markers to the surface of the lung for registration. 
The CT scan parameters were identical to those used in the in vivo experiments.
All experimental steps were performed as was done for the in vivo experiments, with the exception of the use of multiple safe insertion time windows as there is no breathing motion in the ex vivo case.

\subsection*{Overview of the porcine animal model}

We used porcine lungs for our in vivo and ex vivo experiments because of the similarities between the porcine and human lung.
The similarities in anatomical and histological properties make the porcine lung a good translational model for respiratory medicine 
\cite{Judge2014_JRCMB}.
Although porcine lungs share many properties with human lungs, there are several notable differences.
In humans, the interlobular septa between secondary pulmonary lobules are incomplete, allowing for collateral ventilation through the pores of Kohn.
The porcine lung does not have these pores, resulting in little to no collateral ventilation between adjacent lobules.
This anatomical difference results in atelectasis forming more readily in the distal lobes, which is especially noticeable when the bronchoscope is obstructing a portion of the airway lumen. 
This effect limits the depth to which we can safely navigate the bronchoscope in the in vivo porcine lung.
This led us to implement the motion planner to favor more proximal piercing sites, but this did not limit our ability to reach the peripheral lung parenchyma.

For our ex vivo experiments, a notable limitation of the ex vivo porcine lung model is in the formation of high intensity interlobular obstacles postmortem.
The interlobular septa, in which pulmonary veins and lymphatic vessels run, fill with blood and lymphatic fluid postmortem, creating anatomical obstacles that would otherwise not exist in a typical in vivo setting.
In humans, these interlobular septa measure roughly 0.1mm in thickness \cite{Kang1996_JTI}. 
Given this limitation of the ex vivo model, we excluded deployments where the needle path intersected with unsegmented septa greater than 1 mm. 
\subsection*{Statistical analysis}

Statistical analysis was performed in Python using the Scipy package (version 1.7.0).
Unpaired t-tests were used for comparing the values for the lengths and errors in the ex vivo experiments comparing the performance of the robot system and the manual clinical bronchoscopy.
We consider \textit{p} $\lt$ \unit[0.05] to be statistically significant.

\vspace{5mm}
\noindent

\bibliography{bibabbrv, main, method}

\bibliographystyle{IEEEtran}

\section*{Acknowledgments}

\hspace{\parindent}
\textbf{Funding:}
This research was supported in part by the U.S. National Institutes of Health (NIH) under award R01EB024864 and the U.S. National Science Foundation (NSF) under awards 2008475 and 2038855. The content is solely the responsibility of the authors and does not necessarily represent the official views of NIH or NSF.

\textbf{Author contributions:}
R.A.\ and R.J.W.\ conceived the project and oversaw all aspects of the research described herein. 
All six student authors, M.E., T.E.E., I.F., M.F., J.H., and M.R., contributed equally to the overall project. 
A.K., M.E., T.E.E., I.F., M.F., J.H., M.R., J.A, and Y.L.\ conducted experiments. 
A.K., M.E., T.E.E., I.F., M.F., J.H., and M.R.\ analyzed the data.
T.E.E., M.E., and M.R.\ designed and implemented the controller.
M.E., T.E.E., and M.R.\ designed and built the actuation unit, fabricated fiducials, and designed and fabricated the aiming device and steerable needle.
M.F.\ designed and implemented the segmentation algorithm.
A.K., M.F., and J.H.\ designed and implemented the motion planning algorithm.
I.F., M.F. and J.H.\ designed and implemented the visualization software.
M.E., T.E.E., I.F., M.F., J.H., and M.R.\ implemented the registration software.
J.A., E.A.G., Y.L., and F.M.\ provided clinical insight in the development of the system and the experimental methods.
A.K., M.E., T.E.E., I.F., M.F., J.H., M.R., R.J.W., and R.A.\ wrote the manuscript. All other authors performed critical revisions.
The authors thank Gregory Grandio, M.D. for his contribution to the manual bronchoscopy experiments.

\textbf{Competing interests:}
A.K., R.J.W., and R.A.\ are inventors on university-owned patents on medical robotic devices incorporating steerable instruments. The other authors declare no competing interests.

\textbf{Data and materials availability:}
The data that support the findings of this study are provided in the main text of this paper. More data are available from the corresponding author upon reasonable request.
The source code for the steerable needle motion planner is freely available at the following link: 

\url{https://github.com/UNC-Robotics/steerable-needle-planner}.

\newpage

\end{document}